\documentclass{sig-alternate-05-2015}
\usepackage{graphicx}
\usepackage{amsmath,amssymb} 
\usepackage{bm}
\usepackage{mathrsfs}
\usepackage{color}
\usepackage{subfigure}
\usepackage{wasysym}
\usepackage{xcolor}
\usepackage{colortbl,booktabs}
\usepackage{threeparttable}
\usepackage{multirow}
\usepackage{multicol}
\usepackage{wrapfig}

\begin{document}

\setcopyright{acmcopyright}

\doi{ }

\isbn{ }

\conferenceinfo{ICIAI 2019}{March 15-18, 2019, Suzhou, China}

\acmPrice{\$15.00}

%

\title{An Element Sensitive Saliency Model with Position Prior Learning for Web Pages}

%
%
%
%
%

\numberofauthors{4} 
%
\author{
%
%
\alignauthor
Yujun Gu\\
       \affaddr{Cooperative Medianet Innovation Center,}\\
       \affaddr{Shanghai Jiao Tong University, China}\\
       \email{yjgu@sjtu.edu.cn}
\alignauthor
Jie Chang\\
       \affaddr{Cooperative Medianet Innovation Center,}\\
       \affaddr{Shanghai Jiao Tong University, China}\\
       \email{j\_chang@sjtu.edu.cn}
\and
\alignauthor
Ya Zhang\\
       \affaddr{Cooperative Medianet Innovation Center,}\\
       \affaddr{Shanghai Jiao Tong University, China}\\
       \email{ya\_zhang@sjtu.edu.cn}
\alignauthor
Yanfeng Wang\\
       \affaddr{Cooperative Medianet Innovation Center,}\\
       \affaddr{Shanghai Jiao Tong University, China}\\
       \email{yfwang@sjtu.edu.cn}
}


\maketitle
\begin{abstract}
Understanding human visual attention is important for multimedia applications. Many studies have attempted to build saliency prediction models on natural images. However, limited efforts have been devoted to saliency prediction for Web pages, which are characterized by diverse content elements and spatial layouts. In this paper, we propose a novel end-to-end deep generative saliency model for Web pages. To capture position biases introduced by page layouts, a Position Prior Learning (PPL) sub-network is proposed, which models the position biases with a variational auto-encoder. To model different elements of a Web page, a Multi Discriminative Region Detection (MDRD) branch and a Text Region Detection (TRD) branch are introduced, which extract discriminative localizations and prominent text regions, respectively. We validate the proposed model with a public Web-page dataset `FIWI', and show that the proposed model outperforms the state-of-art models for Web-page saliency prediction.

\end{abstract}

%
%


%
%

%
%


\begin{figure}[t]
\centering
\includegraphics[width=\linewidth]{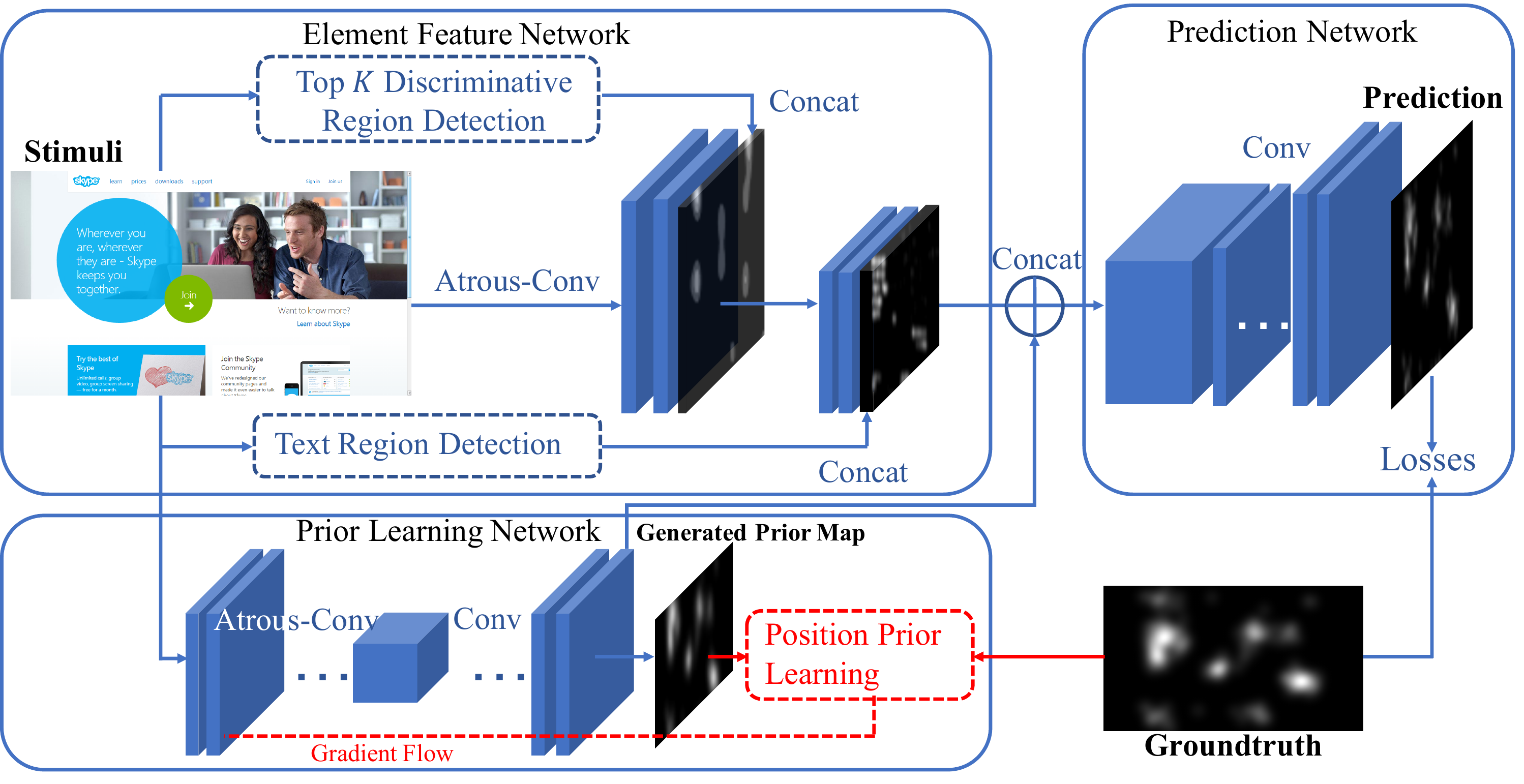}
\caption{An overview of Element Sensitive Saliency Model with Position Prior Learning.}
\label{fig:overview_model}
\end{figure}
\section{Introduction}
\label{introduction}
Dominated by the ``bottom-up'' attentive mechanism of visual cognition~\cite{rensink2000dynamic}, human vision system tends to focus on certain regions instead of randomly spreading. Modeling this human visual attention is essential for evaluating media designs. Inspired by the above visual attention mechanism, many computational \textit{saliency models}, which attempt to predict salient regions of given media content, have been investigated.

Most existing saliency prediction studies focus on natural images~\cite{mit-saliency-benchmark}. Building upon the biological evidences~\cite{zhang2012neural}, low-level features such as color, contrast, luminance, edge orientations or intensity are adopted to help predict human attention~\cite{Itti2000A,harel2007graph}.
To capture the influence of content semantics, high-level features representing certain semantic concepts (e.g., faces, objects) are leveraged to further improve the prediction accuracy~\cite{judd2009learning}.
With the recent development of deep neural network, many efforts have been made to simultaneously learn feature representations and saliency prediction models~\cite{Pan2016Shallow,shen2014learning,cornia2016predicting,cornia2016deep}. 
More recently, adversarial training is leveraged to refine the predictive results of saliency model~\cite{pan2017salgan}.

While much efforts have been devoted to saliency prediction of natural images, there have been very limited studies focusing on Web-page saliency~\cite{shen2014webpage,li2016webpage}.
Different from natural images, Web pages are rich in scattered salient stimuli (e.g., logos, text, graphs, picture)~\cite{still2010saliency} of un-equivalent influence to human's short-term attention~\cite{buscher2009you}. It is thus more difficult to model human attention on Web pages, not only requiring more complicated feature representations but also modeling spatial layouts.
Existing studies on Web-page saliency~\cite{shen2014webpage,li2016webpage} mainly focus on exploring a better feature representation. However, they did not take the characteristics of Web-page saliency into consideration.

First, layout of Web pages greatly affect the deployment of human fixations, leading to a diverse set of reading patterns such as~\cite{buscher2009you}. The above studies for Web-page saliency tend to represent the \emph{position-based visual preferences} using manually constructed position-bias maps, which is unable to adaptively reflect the accurate Web-page layout. Hence, we explore to automatically model the position biases as a prior distribution with the help of a variational auto-encoder.

Second, different from natural images, there are many non-semantic elements in Web pages which may not grip human attention but unavoidably cause overmuch activated regions when we simple use a pre-trained CNN as a feature extractor, just as most previous works done. Instead, considering text and images are dominant elements in Web pages, we propose to adopt independent high-level semantic features for the two elements.

In our paper, we propose a deep generative saliency model for Web pages. As shown in Figure \ref{fig:overview_model}, the whole model consists of three sub-networks: Prior Learning Net (PL-Net) for modeling position biases,  Element Feature Net (EF-Net) for extracting representations for different elements, and Prediction Net(P-Net) for generating the final saliency map.
The PL-Net leverages a VAE-based Position Prior Learning (PPL) algorithm to automatically learn position biases of user viewing behaviors.
The EF-Net contains three branches. In addition to an overall feature branch, a Multi Discriminative Region Detection (MDRD) branch and a Text Region Detection (TRD) branch are introduced, which extract discriminative localizations and prominent text regions, respectively. The whole model we proposed is a deep generative model which can be trained end-to-end. By experimenting on FiWI, a released Web-page dataset, our proposed algorithms distinguish our model and boost the performance of saliency prediction.

The main contributions of our studies are summarized as follows.
\begin{itemize}
\item We model the diverse \textit{visual preference} caused by page layouts with a VAE-based Position Prior Learning.
\item We explore element-based feature representations and leverage a MDRD branch and a TRD branch to capture the impact of images and text to human attention.
\item Experimental studies have shown that the proposed method outperforms the state-of-art models for Web-page saliency prediction.
\end{itemize}

\section{Method}
\label{method}
Figure~\ref{fig:overview_model} provides an overview of the proposed generative saliency model for Web pages. In the rest of this section, we present PPL, MDRD and TRD in details.

\subsection{Position Prior Learning}
\label{sub: learned-prior-transfer}
\begin{figure}[t]
\centering
\includegraphics[width=0.95\linewidth]{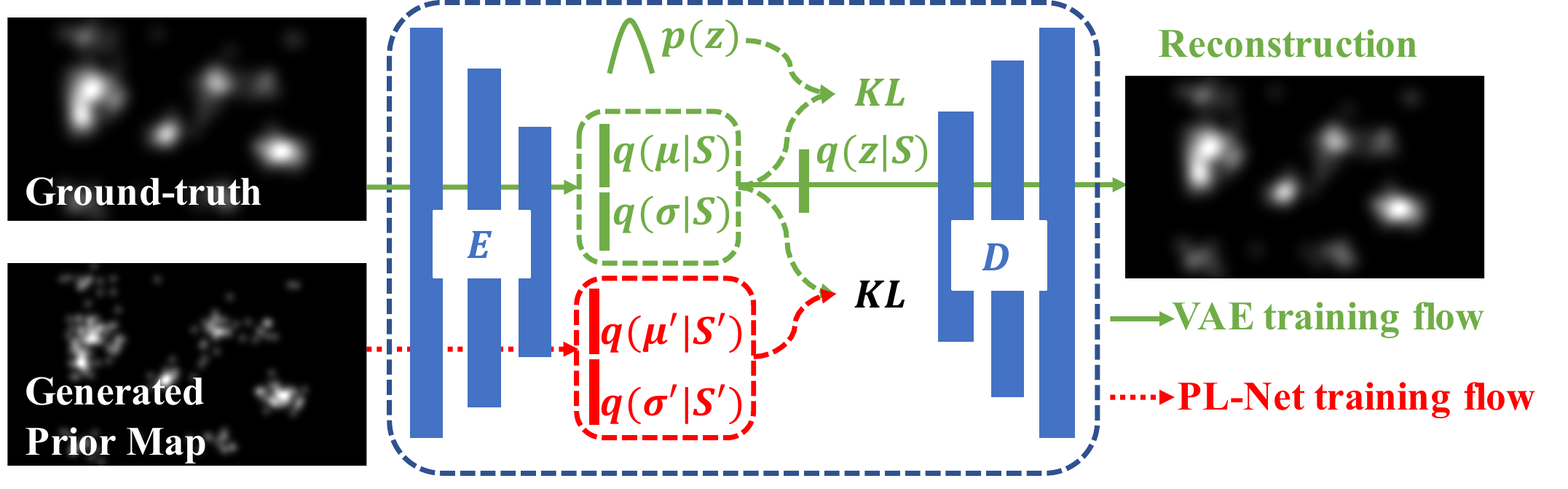}
\caption{The architecture of Position Prior Learning (PPL) sub-network. The corresponding ground-truth of the trained stimulus is reconstructed by a variational auto-encoder for a learned posterior distribution. Meanwhile another latent distribution inferred from the generated prior map of PL-Net is aligned with the previous learned posterior by KL divergence term (marked in black).}
\label{fig: prior_learning}
\end{figure}

To predict Web page saliency, it is important to model position biases introduced by page layouts. Unlike previous studies which adopt fixed position bias maps manually constructed beforehand, we propose a Position Prior Learning (PPL) algorithm based on Variational Auto-Encoder(VAE)~\cite{kingma2013auto} to automatically learn such position biases.
Based on the observation that similar position biases occurred on corresponding Web pages sharing similar layouts. We explore to model these postion biases as the mean and standard deviation of multivariate Gaussian distribution which are learned as latent variables in VAE (see Fig~\ref{fig: prior_learning}).
Specifically, a set of true saliency maps $S$ are be used to optimized a VAE network including an encoder ${E}$ and a decoder ${D}$ for obtaining the posterior distribution $q(z|S)$. This training procedure follows the objective function below:
\begin{equation}
L(\theta,\phi;S) = \lambda_{1}\mathbb{E}_{z\sim q_{\phi}(z|S)}[\log{p_{\theta}(S|z)}] - \lambda_{2}D_{KL}(q_{\phi}(z|S)||p(z)),
\label{eq: eq6}
\end{equation}
where $\phi$ and $\theta$ are respective parameters in encoder $E$ and decoder $D$ of VAE, $p(z)$ is a standard normal prior, $N(\bm{0},\bm{I})$, $\lambda_1$ and $\lambda_2$ control the weights of the expectation term and KL-divergence terms.
The learned variational approximate posterior can be formulated as a multivariate Gaussian with a diagonal covariance structure,
\begin{equation}
q_{\phi}(\bm{z}|S) = N(\bm{z};\bm{\mu},\bm{\sigma}^{2}\bm{I}|S),
\label{eq: eq7}
\end{equation}
where the mean and standard deviation of the approximate posterior, $\bm{\mu^{(i)}}$ and $\bm{\sigma^{(i)}}$, are outputs of the encoding MLP ($E$).

Meanwhile, the generated prior maps $S^{'}$ from PL-Net are fed into the parameter-sharing encoder $E$. We also let the output of $E$ be a multivariate Gaussian structure,
\begin{equation}
q_{\phi}(\bm{z^{'}}|S^{'}) = N(\bm{z}^{'};\bm{\mu}^{'},{\bm{\sigma}^{2}}^{'}\bm{I}|S^{'}),
\label{eq: eq8}
\end{equation}
where $\phi$ are reused parameters of encoder $E$.

Then another KL-divergence representing the discrepancy between the above two approximate posterior, $q_{\phi}(\bm{z}|S)$ and $q_{\phi}(\bm{z^{'}}|S^{'})$,
\begin{equation}
L(\theta_{pl}) = D_{KL}(N(\bm{z}^{'};\bm{\mu}^{'},{\bm{\sigma}^{2}}^{'}\bm{I}|S^{'}) || N(\bm{z};\bm{\mu},\bm{\sigma}^{2}\bm{I}|S))
\label{eq: eq9}
\end{equation}
is calculated as the loss for training PL-Net so as to enable prior maps generated by PL-Net to possess similar latent variables with that of true saliency maps. $\theta_{pl}$ indicates the parameters in PL-Net.

\subsection{Multi Discriminative Region Detection}
\label{sub: multi-discriminative-region-detection}
We propose Multiple Discriminative Region Detection (MDRD) to extract the remarkable object regions where human maybe easily focus on. Inspired by Class Activation Map (CAM) proposed in~\cite{zhou2016learning}, first, we utilize a VGG16-GAP model trained on ImageNet~\cite{krizhevsky2012imagenet} to predict the classification of each input stimuli image. Then we select top-$K$ categories predicted by the model, and calculate the average of top-$k$ categories CAM to get the multi-discriminative region map $M$.
\begin{equation}
\begin{split}
M(x, y) = \frac{1}{K}\sum_{c \in C}S_c(x, y), \\
C = \{{c|Top_K(Y_{c}, c=1,2,..., \#)}\},
\end{split}
\label{eq: eq3}
\end{equation}
where $(x,y)$ is the position index of pixels, $Y_c$ is the probabilistic value w.r.t class $c$; $S_c$ is the CAM of category $c$; $\#$ is the number of all categories in ImageNet; $Top_K$ is the function which return a set of class numbers whose predicted scores are in top-K. $K$ is determined by the number of dominated eigenvalues after PCA implemented on the last convolutional layers.

\subsection{Text Region Detection}
\label{sub: text-region-detection}
Web pages are rich in text information which greatly attracted human fixations, hence our proposed Text Region Detection (TRD) aims to generate representation for prominent text information.

TRD is mainly implemented by a Text/Background Classifier $C_f$ trained on the datasets (ICDAR~\cite{ICDAR2003} and SVT~\cite{wang2010word}) for character recognition. A well-trained $C_f$ is then performed around the resized multi-scale input stimulus by sliding window for generating the text saliency map. Guassian blur is applied to smooth the text saliency map.
%

\subsection{Loss Function}
\label{sub: loss-function}
The feature maps from PL-Net and EF-Net are concatenated together as the input of Prediction Network (P-Net). P-Net generates the final predicted saliency map based on stacked CNN structure. The loss function we defined between the predicted saliency map and its corresponding ground-truth is the linear combination of two terms as follows:
\begin{equation}
L(\theta^{*}) = \alpha L_1(\hat{S},S) + \beta L_2(\hat{S}, S),
\label{eq: eq10}
\end{equation}
where $\theta^{*}$ are training parameters in EF-Net and P-Net; $\hat{S}$ are predicted saliency maps from P-Net and $S$ are corresponding ground-truth; $\alpha$, $\beta$ are hyper-parameters to trade-off two loss terms where
$L_1$ is defined as the cross entropy loss:
\begin{equation}
L_1(\hat{S},S) = \mathbb{E}_{s,\hat{s}}[S\log(\hat{S}) - (1-S)\log(1-\hat{S})],
\label{eq: eq11}
\end{equation}
and $L_2$ is defined as the KL-divergence measuring the loss of information when distribution $\hat{S}$ is used to approximate the distribution $S$:
\begin{equation}
L_2(\hat{S},S) = \sum_{i}\hat{S_i}\log(\frac{\hat{S_i}}{S_{i}+\varepsilon}+\varepsilon),
\label{eq: eq12}
\end{equation}
where $i$ indicates the $i^{th}$ pixel in both saliency maps and $\varepsilon$ is a regularization constant.

\section{Datasets \& Metrics}
\textbf{FiWI} is a dataset proposed in~\cite{shen2014webpage}, which contains 149 Web page screenshots with eye-tracking fixation data collected from 11 observers. The observation is short-term as well as free-viewing to ensure the visual preference being driven by ``bottom-up'' visual mechanism.
FiWI is categorized as Pictorial(50), Textual(50) and Mixed(49) images according to the different composition of text and pictures. Pictorial Web pages are occupied by pictures and less text, Textual Web pages contains informative text with high density and Mixed Web pages are a mix of pictures and text.


\textbf{Evaluation Metrics} For evaluating our performance quantitatively, three similarity metrics\footnote{https://sites.google.com/site/saliencyevaluation/evaluation-measures} are adopted including Linear Correlation Coefficient (CC), Normalized Scanpath Saliency (NSS) and shuffled Area Under Curve (sAUC)~\cite{zhang2008sun}.


\label{experiments-evaluation}
\begin{figure}[]
\centering
\includegraphics[width=0.95\linewidth]{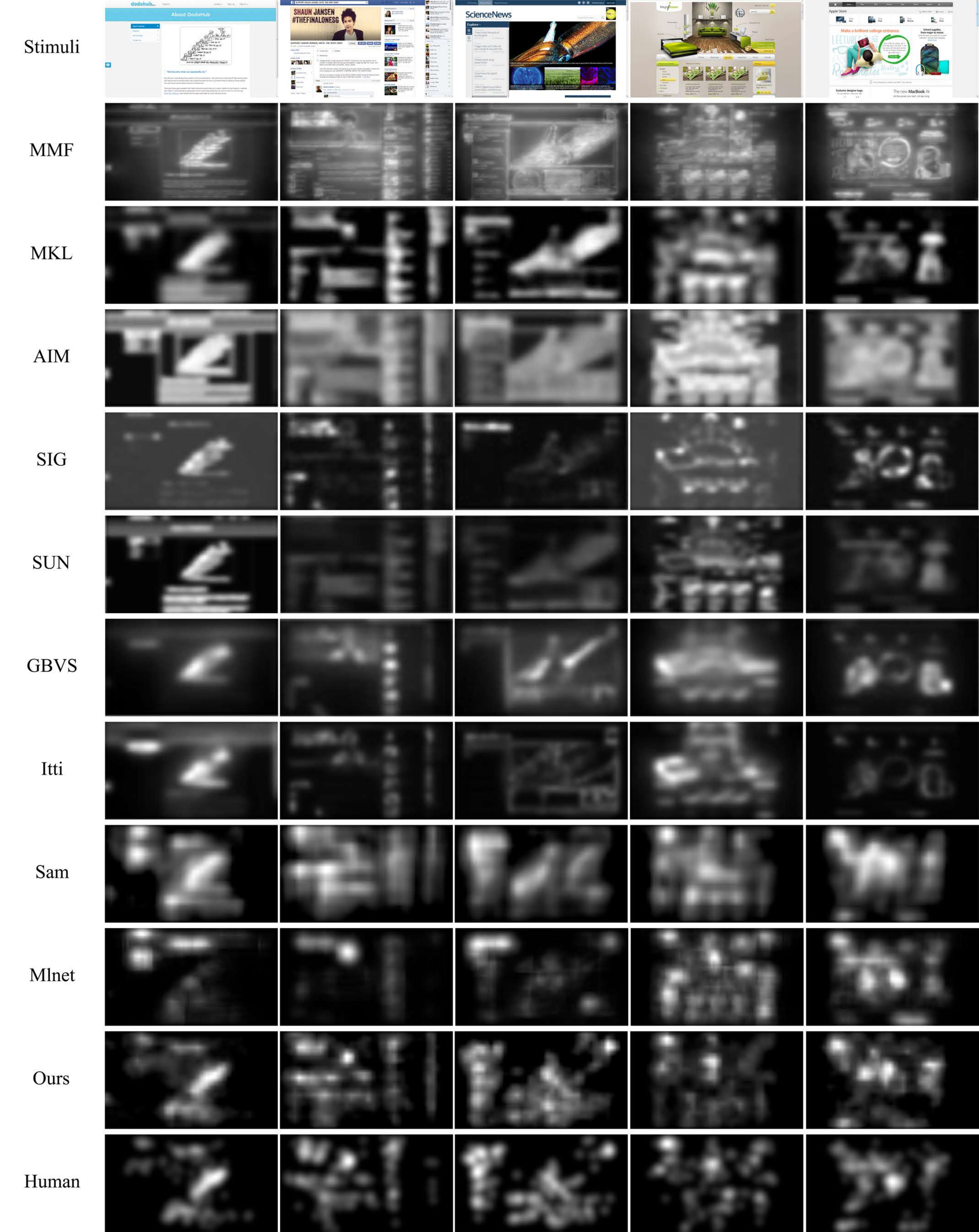}
\caption{Qualitative results and comparison to the state of the art. Compared with MMF and MKL without considering ``promident'' text information, our predicted saliency maps have more accurate response at textual location. We also outperform other baselines (AIM$\sim$Mlnet) proposed for natural images by avoiding patches of irrelevant high-response.}
\label{fig: performance_comparison}
\end{figure}

\section{Experiments Results}

In this section, we first qualitatively evaluate our model with existing nine studies proposed for saliency prediction. We also present a quantitative comparison on Pictorial, Text and Mixed images from FiWI. Furthermore, we analyze the effectiveness of each component of the proposed algorithm by removing position prior learning (PPL), multi-discriminative region detection (MDRD) and text region detection (TRD) from the whole network. Last, we experimentally verifies that the proposed TRD and MDRD can be plugged into other saliency models and we show that they lead to performance gains on top of two state-of-art saliency models, Sam~\cite{cornia2016predicting} and Mlnet~\cite{cornia2016deep}.

\subsection{Performance Comparison}
\label{sub: performance-comparison}

We compare our model with nine previous saliency models, including: MMF~\cite{li2016webpage}, MKL~\cite{shen2014webpage}, AIM~\cite{bruce2009saliency}, SIG~\cite{hou2012image}, SUN~\cite{zhang2008sun}, GBVS~\cite{harel2007graph}, Itti~\cite{Itti2000A}, Sam~\cite{cornia2016predicting} and Mlnet~\cite{cornia2016deep}. Figure~\ref{fig: performance_comparison} illustrates the comparison results among the models, which demonstrates that our model better represents human attention.
In Table~\ref{tab: QuantitativeMeasurements1}, we quantitatively compare the performance in terms of three evaluation metrics (sAUC, NSS and CC) for Pictorial/Text/Mixed Web pages.
It can be seen that our model greatly outperforms other baselines in Pictorial\&Text Web pages and is slightly better in Mixed Web pages.

\begin{table*}[]
\centering
\caption{Quantitative Measurements of Different Methods}
\begin{threeparttable}

  \begin{tabular}{lccccccccc}
  \toprule[1pt]
  \multirow{2}{*}{Model}&
  \multicolumn{3}{c}{Pictorial-webpage}&\multicolumn{3}{c}{Text-webpage}&\multicolumn{3}{c}{Mixed-webpage}\\
  \cmidrule(lr){2-4} \cmidrule(lr){5-7} \cmidrule(lr){8-10}
  &sAUC&NSS&CC&sAUC&NSS&CC&sAUC&NSS&CC\\
  \midrule
  Itti~\cite{Itti2000A}  &0.538&0.443&0.233 &0.544&0.483&0.234 &0.552&0.462&0.267\\

  \rowcolor[gray]{0.9} GBVS~\cite{harel2007graph} &0.564&0.685&0.350 &0.561&0.680&0.329 &0.575&0.708&0.336\\

  AWS~\cite{garcia2012relationship}  &0.633&0.806&0.401 &0.648&0.867&0.406 &0.645&0.854&0.393 \\

  \rowcolor[gray]{0.9} Signat~\cite{hou2012image} &0.664&0.848&0.415 &0.682&0.885&0.408 &0.682&0.920&0.410\\

  AIM~\cite{bruce2009saliency}  &0.663&0.907&0.453 &0.679&0.936&0.445 &0.678&0.943&0.439\\

  \rowcolor[gray]{0.9}SUN~\cite{zhang2008sun} &0.707&1.072&0.488 &0.687&0.993&0.459 &0.700&1.017&0.490\\

  MKL~\cite{shen2014webpage} &0.723&0.880&0.429 &0.741&0.861&0.410 &0.730&0.891&0.433\\

  \rowcolor[gray]{0.9}MMF~\cite{li2016webpage} &0.731&0.904&0.441 &0.720&0.890&0.419 &0.760&0.920&0.431\\

  Mlnet~\cite{cornia2016deep} &0.703&0.912&0.530 &0.711&0.802&0.463 &0.725&0.905&0.522\\

  \rowcolor[gray]{0.9}Sam~\cite{cornia2016predicting} &0.720&0.982&0.494 &0.743&0.924&0.470 &\textbf{0.762}&0.938&0.500\\

  Ours &\textbf{0.761}&\textbf{1.202}&\textbf{0.641} &\textbf{0.781}&\textbf{1.250}&\textbf{0.511} &0.751&\textbf{1.029}&\textbf{0.580}\\

  \bottomrule[1pt]
  \end{tabular}
\label{tab: QuantitativeMeasurements1}
\end{threeparttable}
\end{table*}

\subsection{Analysis of Each Module}
\label{sub: analysis-of-each-module}
\begin{figure}[t]
\centering
\includegraphics[width=\linewidth]{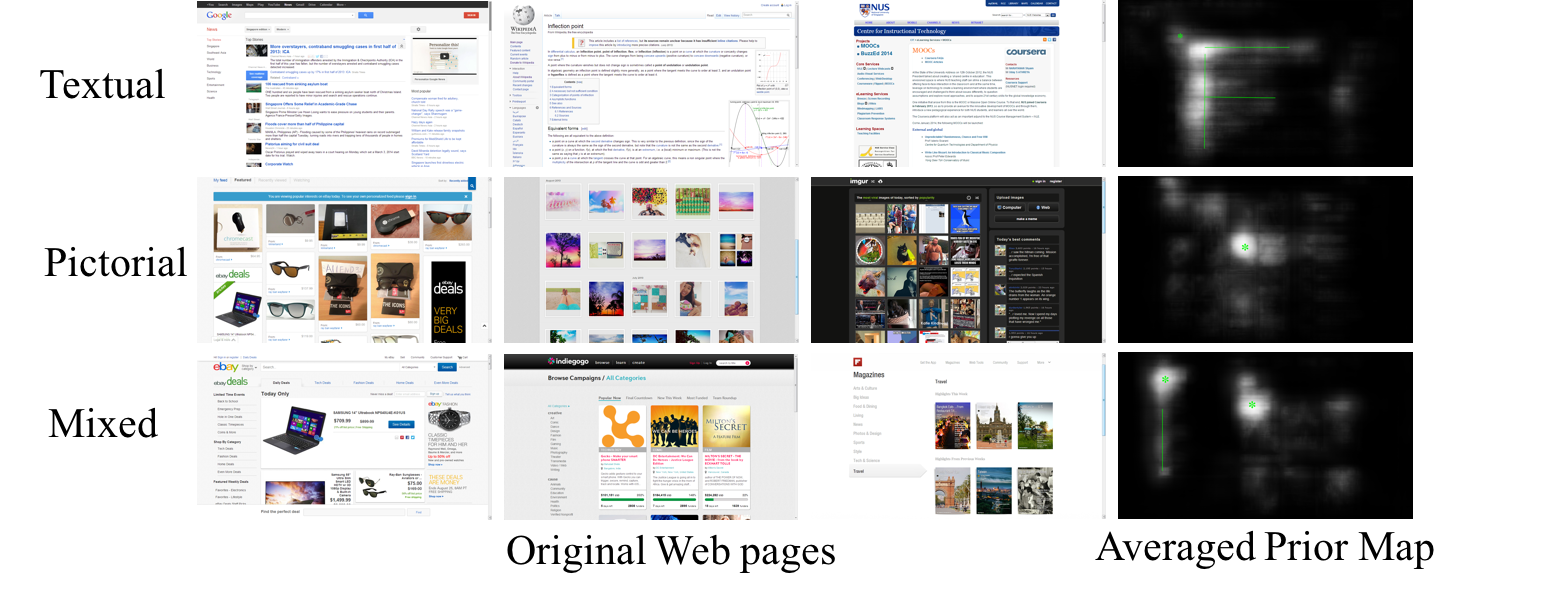}
\caption{Visualization of the typical position biases learned by Prior Learning Network. Last column shows the corresponding averaged results generated from Prior Learning Network for each kind of Web pages.}
\label{fig: position-bias}
\end{figure}

We further analyze the respective effect of PPL, MDRD and TRD proposed in our Element Sensitive Saliency Model.
First, to explore whether Position Prior Learning captures position bias in web page viewing, we illustrate in Fig~\ref{fig: position-bias} with three kinds of Web pages: web pages rich in text information, web pages arranged by pictures, and web pages combined with pictures and text. For each category, we average their corresponding prior maps $S^{'}$ generated by Prior Learning Network and we observe typical ``F-shaped'' and ``top-left'' bias in textual web pages; ``center-arround'' bias in pictorial images; ``sidebar'' and ``top-left'' bias in mixed web pages. That means the proposed PPL algorithm is able to capture common prior of position bias for Web pages with similar layouts.

\begin{figure}[t]
\centering
\includegraphics[width=\linewidth]{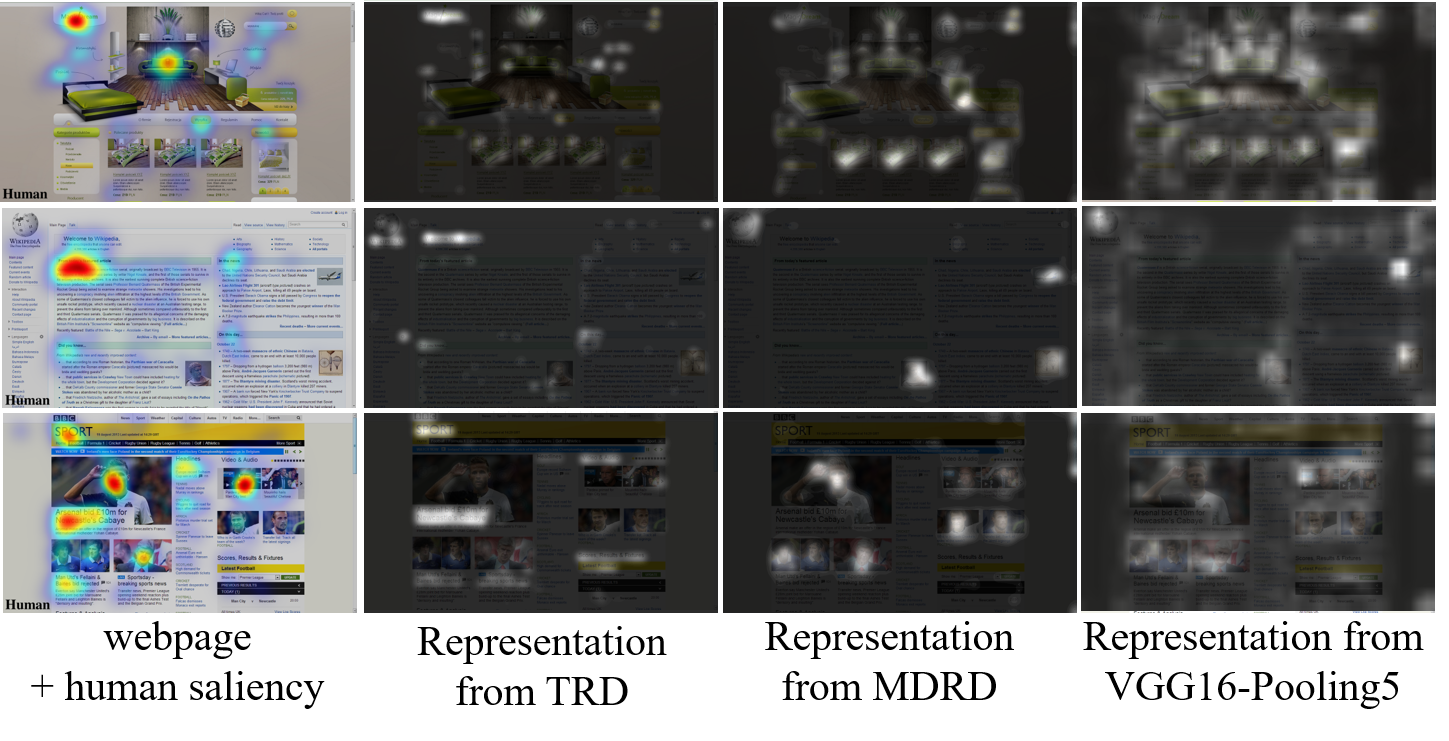}
\caption{Visualization of what TRD and MDRD extracted.}
\label{fig: visualize}
\end{figure}

Then we intuitively visualize what TRD and MDRD extracted from original Web pages in figure~\ref{fig: visualize}. Representation from TRD shows that TRD selectively highlights locations where textual information is remarkable instead of simply detecting the edge lines of each character used in previous method. We see text in logos, headlines or subheadings have larger activation on text saliency maps, which is important for our model since human usually pay more attention on these regions than normal text in main bodies. Representation from MDRD also shows that the proposed MDRD could ``pre-select'' some special discriminative regions while suppress textual regions greatly. For comparison, as most previous works done, representation from pooling5 is extracted from pre-trained VGG16~\cite{simonyan2014very}, which shows feature maps generated by our MDRD are more sparse with most inconspicuous regions suppressed.


Furthermore, we quantitatively illustrate the effectiveness of TRD / MDRD / PPL modules. Table~\ref{tab: QuantitativeMeasurements2}  compares  ``Base'', ``Base+TRD'', ``Base+MDRD'', ``Base+TRD+MDRD'' and the proposed model, ``Base+TRD+MDRD+PPL'', in terms of sAUC, NSS and CC metrics, which shows each module greatly contributes to saliency prediction for Web pages.
\begin{table}[]
\centering
\caption{Ablation Study}
\begin{threeparttable}

  \begin{tabular}{lccc}
  \toprule[1pt]
  \multirow{2}{*}{Prunning Experiment}&
  \multicolumn{3}{c}{All Test Webpages}\\
  \cmidrule(lr){2-4}
    &sAUC&NSS&CC\\
  \midrule
  Baseline  &0.545&0.561&0.402\\

  \rowcolor[gray]{0.9} Baseline+TRD &0.730&0.855&0.511 \\

  Baseline+MDRD  &0.700&0.820&0.495  \\

  \rowcolor[gray]{0.9} Baseline+TRD+MDRD &0.731&1.005&0.612\\

  Baseline+TRD+MDRD+PPL  &\textbf{0.760}&\textbf{1.085}&\textbf{0.637}\\

  \bottomrule[1pt]
  \end{tabular}
\label{tab: QuantitativeMeasurements2}
\end{threeparttable}
\end{table}


\section{Conclusion}
\label{conclusion-and-future-work}
In this paper, we present a Element Sensitive Saliency Model for Web pages. The whole model consists of Element Feature Network (EF-Net), Prior Learning Network (PL-Net) and Prediction Network (P-Net).
Compared with previous works, we propose VAE-based Position Prior Learning in PL-Net to automatically learn the various visual preference when human scan Web pages. Additionally, in EF-Net, we leverage Text Region Detection(TRD) and Multi-Discriminative Region Detection(MDRD) to handle specific challenges in this task. We experimentally verified that the proposed model outperforms the state-of-art models for Web-page saliency prediction.

%
\bibliographystyle{abbrv}
\bibliography{egbib}  

\clearpage
\begin{table*}[]
\caption{Authors' background}
\begin{tabular}{|l|l|l|l|l}
\cline{1-4}
Yujun Gu     & master student & Computer Vision                &                                      &  \\ \cline{1-4}
Jie Chang    & master student & Computer Vision                &                                      &  \\ \cline{1-4}
Ya Zhang     & full professor & Computer Vision \& Data Mining & http://ir.sjtu.edu.cn/$\sim$yazhang/ &  \\ \cline{1-4}
Yanfeng Wang & full professor & Computer Vision                &                                      &  \\ \cline{1-4}
\end{tabular}
\end{table*}

\end{document}